\documentclass[10pt,twocolumn,letterpaper]{article}

\usepackage[accsupp]{axessibility} 
\usepackage{cvpr}              
\definecolor{cvprblue}{rgb}{0.21,0.49,0.74}
\usepackage[pagebackref,breaklinks,colorlinks,allcolors=cvprblue]{hyperref}

\def\paperID{45818} 
\def\confName{CVPR}
\def\confYear{2026}

\title{Breaking Smooth-Motion Assumptions: A UAV Benchmark for Multi-Object Tracking in Complex and Adverse Conditions}

\author{
	Jingtao Ye\thanks{Equal contribution},\hspace{1mm} Kexin Zhang\footnotemark[1],\hspace{1mm} Xunchi Ma,\hspace{1mm} Yuehan Li\\
	Guangming Zhu,\hspace{1mm} Peiyi Shen\textsuperscript{\ddag},\hspace{1mm} Linhua Jiang,\hspace{1mm} Xiangdong Zhang,\hspace{1mm} Liang Zhang\textsuperscript{\ddag},\hspace{1mm}\\
	Xidian University, School of Computer Science and Technology, China \\
    \textsuperscript{*} Equal contribution, \hspace{1mm} \textsuperscript{\ddag} Corresponding Authors
}

\usepackage{enumitem}
\usepackage{multirow}
\usepackage{algorithm}
\usepackage{algpseudocode}
\usepackage{hyperref}

\begin{document}
\maketitle
\begin{abstract}
The rapid movements and agile maneuvers of unmanned aerial vehicles (UAVs) induce significant observational challenges for multi-object tracking (MOT). However, existing UAV-perspective MOT benchmarks often lack these complexities, featuring predominantly predictable camera dynamics and linear motion patterns. To address this gap, we introduce DynUAV, a new benchmark for dynamic UAV-perspective MOT, characterized by intense ego-motion and the resulting complex apparent trajectories. The benchmark comprises 42 video sequences with over 1.7 million bounding box annotations, covering vehicles, pedestrians, and specialized industrial categories such as excavators, bulldozers and cranes. Compared to existing benchmarks, DynUAV introduces substantial challenges arising from ego-motion, including drastic scale changes and viewpoint changes, as well as motion blur. Comprehensive evaluations of state-of-the-art trackers on DynUAV reveal their limitations, particularly in managing the intertwined challenges of detection and association under such dynamic conditions, thereby establishing DynUAV as a rigorous benchmark. We anticipate that DynUAV will serve as a demanding testbed to spur progress in real-world UAV-perspective MOT, and we will make all resources available at \href{https://github.com/kxzhang-lab/DynUAV/tree/master}{link}.
\end{abstract}    
\section{Introduction}
\label{sec:intro}

\noindent MOT has made significant progress in surveillance and autonomous driving applications\cite{guo2022review}, but UAVs present distinct challenges. Compared to fixed surveillance cameras or vehicle-mounted systems, the UAV-perspective introduces unique difficulties for MOT, including small object sizes, complex motion patterns, and diverse environmental conditions\cite{bashar2022multiple}. While existing UAV-perspective MOT benchmarks address challenges such as occlusion, illumination changes, scale variations, and viewpoint differences, they still exhibit the following limitations:

\par\noindent  
\begin{itemize}[leftmargin=*]
	\item UAV-perspective trajectories within a single video are often straightforward, producing smooth and nearly linear object paths in the image plane.
	\item The scene diversity is limited, as most sequences are confined to conventional urban streets and intersections. Furthermore, the object categories are typically restricted to common road vehicles like cars and buses.
\end{itemize}

However, in real-world applications, UAVs frequently execute dynamic, nonlinear, and rapid maneuvers\cite{telli2023comprehensive}, which impose severe observational challenges on MOT. For instance, in public safety and defense, UAVs must often maneuver flexibly to avoid obstacles in complex urban environments\cite{mohsan2023unmanned}. In urban traffic monitoring, UAVs perform lateral flights, ascents, and descents to effectively capture broad multi-lane scenarios\cite{nagrare2024intersection}. For sports and crowd analysis, UAVs often accelerate or decelerate to track key targets, inducing drastic scale variations and motion blur\cite{zachariadis20172d}. These canonical examples illustrate that complex motion is inherent to UAV maneuvers, yet current benchmarks still provide markedly insufficient coverage for these highly dynamic operational conditions.

To address this gap, we introduce DynUAV, a large-scale UAV-perspective tracking benchmark. It is specifically designed not merely to record dynamic objects, but to capture the challenges arising from severe ego-motion, which is induced by deliberately agile UAV maneuvers.

\begin{itemize}[leftmargin=*]
	\item DynUAV features complex trajectories and significant motion blur, deliberately breaking the smooth-motion assumption that underpins existing MOT algorithms.
	\item Our benchmark spans a wide array of highly structured environments, from urban traffic and campuses to industrial sites. This unprecedented diversity provides an ideal platform for evaluating and rigorously improving the generalization capabilities of tracking models across varied operational domains.
	\item According to our statistics, DynUAV has the longest average sequence duration among comparable UAV-perspective MOT benchmarks, which increases the challenges of temporal modeling and error accumulation issues that are critical in extended tracking scenarios.
\end{itemize}

\section{Related Works}
\label{sec:rel}

\noindent\textbf{Established MOT benchmarks.} Foundational benchmarks like KITTI\cite{geiger2013vision}, UA-DETRAC\cite{wen2020ua}, and MOT Challenge rely on static or ground-level viewpoints. While pivotal, these perspectives inherently limit trajectory complexity, missing the non-linear visual shifts caused by 3D camera movement. Recent works like DanceTrack\cite{sun2022dancetrack} and SportsMOT\cite{cui2023sportsmot} challenge the smooth-motion assumption by introducing rapid movements of \textit{targets}. However, their primary challenge stems from the complex behavior of objects rather than the sensor's ego-motion, leaving a gap in evaluating robustness against camera-induced dynamics common in aerial robotics.

\noindent\textbf{UAV-perspective MOT benchmarks.} UAV benchmarks are categorized into multi-purpose or task-specific. Multi-purpose datasets like UAVDT\cite{du2018unmanned} and VisDrone\cite{zhu2021detection} offer diverse attributes but are predominantly captured during high-altitude hovering or slow cruising, resulting in relatively smooth camera trajectories. 
Task-specific benchmarks address distinct challenges. MDMT\cite{liu2023robust} and \textbf{M3OT\cite{nie2025m3ot}} focus on multi-drone or multi-modality fusion, prioritizing cross-view cues. \textbf{SMOT4SB}\cite{kondo2025mva} targets the flocking of indistinguishable small objects (e.g., birds), where appearance-based re-identification is effectively absent. NAT2021\cite{ye2022unsupervised} targets nighttime tracking. 
Regarding motion, DBT70\cite{li2017visual} uses predefined flight paths, while \textbf{BioDrone}\cite{zhao2024biodrone} highlights frame-to-frame disruption from UAV shake. Crucially, BioDrone\cite{zhao2024biodrone} is a Single Object Tracking (SOT) benchmark, lacking the complex data association challenges inherent to multi-object scenarios. Other datasets like UAV123\cite{benchmark2016benchmark} remain popular for generic evaluation.

\noindent\textbf{Summary.} While existing benchmarks have driven progress, they are limited by fixed viewpoints, simplistic flight patterns\cite{ding2025vision}, or specific task constraints (e.g., SOT, swarms). To bridge this gap, DynUAV systematically captures the severe ego-motion of agile UAV flights, introducing diverse challenging patterns essential for advancing general-category MOT research.

\section{DynUAV Dataset}
\label{sec:statis}

\subsection{Dataset Construction}

\noindent\textbf{Dataset design.} MOT in real-world UAV applications faces fundamental challenges due to agile maneuvers that cause rapid motion and abrupt viewpoint shifts. To address the limitations of existing benchmarks, which assume smooth motion, we introduce DynUAV, a benchmark that \emph{actively leverages UAV maneuverability to simulate highly dynamic scenarios}. It \emph{systematically incorporates complex apparent trajectories and frequent rapid motion events through controlled velocity changes and flexible camera poses}.
Beyond dynamic content, our design incorporates specific features essential for advancing MOT research: (1) Long-term robustness: DynUAV sequences have longer average durations compared to similar datasets, requiring trackers to maintain object identities over extended periods and effectively address the significant challenge of error accumulation. (2) Ego-motion disentanglement: The design includes sequences featuring both complex UAV movements and static objects. This comprehensive approach directly challenges current MOT paradigms by pushing models to move beyond simplistic motion assumptions and develop enhanced temporal reasoning capabilities to tackle critical issues such as motion blur and identity preservation under realistic, highly dynamic conditions.

\noindent\textbf{Video collection.} To fulfill our design objectives, we employed a diverse set of flight strategies\textemdash \emph{including stationary hovering, cruising, variable-speed motion, rotation, and zooming}\textemdash to simulate realistic observational challenges such as viewpoint shifts, scale variation, motion blur, and occlusion. All data were captured using a DJI Mini 4 Pro, equipped with a 1/1.3-inch CMOS sensor at 1080p resolution, under varied weather and lighting conditions. Flights were conducted at altitudes of $80m$ to $120m$ and incorporated orbital and fly-in/pull-away maneuvers to intentionally introduce severe ego-motion. Furthermore, the dataset includes extensive nighttime sequences to enhance its spatio-temporal diversity.

\begin{figure*}[t!]
	\centering
	\includegraphics[width=1.0\textwidth]{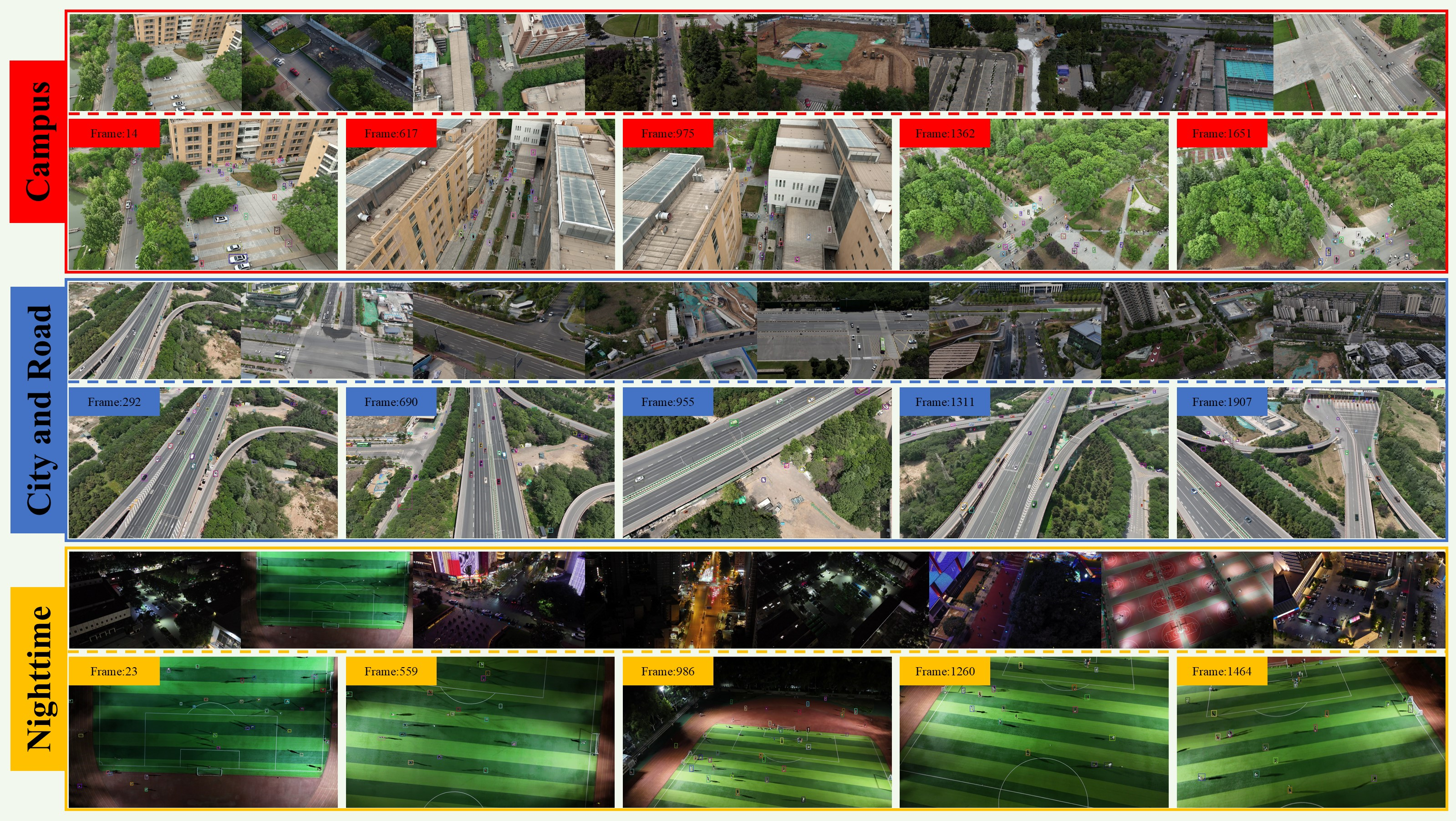} 
	\captionsetup{width=1.0\textwidth, justification=justified}
	\caption{Visualization of scene diversity and complexity in the DynUAV benchmark. Our benchmark encompasses three primary scenarios: campus, city and road, and nighttime. For each scenario, the upper montage depicts a variety of its constituent sequences, while the lower panel illustrates the temporal progression of a single representative sequence, accompanied with its ground-truth annotations.}
	\label{fig:Dataset scene display}
    \vspace{-4mm}
\end{figure*}

\noindent\textbf{Annotation.} All sequences were annotated on the CVAT platform, with each object labeled with its bounding-box coordinates, category, frame index, and a unique track ID. Our annotation protocol dictates that a target is labeled only from the frame it first becomes fully visible. During occlusions, the object's position is continuously annotated as long as its location can be reliably inferred from trajectory context. To guarantee annotation quality, we implemented a rigorous three-stage pipeline: Initial, Review, and Refinement. The initial stage involved semi-automatic labeling, which was then followed by a peer-review process for verification. In the final refinement stage, we employed a custom visualization script to automate error detection via both frame-level and ID-level sampling. Guided by this tool, annotators adhered to strict principles of bounding-box precision and temporal identity consistency, yielding highly reliable ground-truth trajectories.

\noindent\textbf{Summary.} The DynUAV benchmark was meticulously designed with comprehensive considerations of scene diversity, flight dynamics, and annotation accuracy. By incorporating varied lighting conditions, multi-perspective flight strategies, and a rigorous multi-stage labeling pipeline, DynUAV achieves high realism and complexity across spatial, temporal, and categorical dimensions. However, we acknowledge limitations regarding the dataset scale, constrained by the high cost of manual annotation, and the absence of extreme weather conditions due to flight safety regulations.

\subsection{Dataset Statistics}
This section provides a statistical analysis of the DynUAV benchmark, with a comparative analysis against existing benchmarks presented subsequently.

\noindent\textbf{Scale and splits.} The DynUAV benchmark consists of 42 sequences, 37,893 frames, and over 1.7 million bounding box annotations. We divided the dataset into training (30 sequences), validation (5 sequences), and testing (7 sequences) sets, ensuring a balanced distribution of key attributes such as sequence length, scene type, motion intensity, and object density across all splits.

\noindent\textbf{Scene diversity.} As depicted in Fig.~\ref{fig:Dataset scene display}, DynUAV covers three primary scenes: campus, city and road, and nighttime. A key feature is the introduction of underrepresented industrial vehicles, such as ``excavator", ``crane" and ``bulldozer". The campus scenes, with their dense crowds, are designed to test association capabilities, while the high-speed traffic in road scenarios challenges motion modeling. Furthermore, the nighttime sequences, characterized by varied illumination, facilitate research in cross-domain detection.

\noindent\textbf{Object categories.} DynUAV defines eight object categories across vehicles and pedestrians. \emph{Beyond conventional cars and trucks, we include industrial vehicles like excavators, cranes and bulldozers.} For annotation consistency, \emph{all two-wheeled vehicles are grouped as ``cycle", with their riders labeled as ``cycler".} Additionally, \emph{the ``person" class is used for pedestrians.} This rich categorical diversity provides a solid foundation for analyzing multi-class interactions in complex aerial views.

\subsection{Comparative Analysis}
\label{subsec:statistic compare}

\noindent We quantitatively compare DynUAV with existing UAV-perspective MOT benchmarks in terms of scale, temporal consistency, object size, and trajectory continuity. 

\begin{table*}[t]
	\setlength{\tabcolsep}{4pt} 
	\centering
    \caption{Dataset composition and frame statistics. Overview of sequence length (Seq), bounding boxes (BBox, k=$\times10^3$), task type (UAV/general; S/M/D), and frame statistic (sum, minimum, average, maximum) compared to existing MOT benchmarks.}
	\begin{tabular}{l|c|c|ccccccc}
		\toprule
		\multicolumn{2}{c}{Dataset} & MOT-17\cite{milan2016mot16} & MOT-20\cite{dendorfer2020mot20} & VisDrone\cite{zhu2021detection}
		& UAVDT\cite{du2018unmanned} & MDMT\cite{liu2023robust} & DanceTrack\cite{sun2022dancetrack} & DynUAV \\
		\midrule
		\multicolumn{2}{c}{Task} & General M & General M & UAV M,S,D & UAV M,S,D & UAV M & General M & UAV M \\
		\hline
		\multicolumn{2}{c}{Seq.} & 42 & 8 & 92 & 50 & 88 & \textbf{100} & 42 \\
		\hline
		\multicolumn{2}{c}{BBox} & 614k & 1337k & 1621k & 799k & \textbf{2212k} & 574k & 1720k \\
		\hline
		\multirow{4}{*}{Frame} & Min & 450 & 429 & 58 & 265 & 151 & 183 & \textbf{1076} \\
		\cline{2-9}
		& Avg & 802 & 1676 & 417 & 815 & 451 & 1059 & \textbf{1828} \\
		\cline{2-9}
		& Max & 1500 & 3315 & 1424 & 2035 & 703 & 2402 & \textbf{2655} \\
		\cline{2-9}
		& Sum & 34k & 13k & 40k & 41k & 40k & \textbf{106k} & 38k \\
		\bottomrule
	\end{tabular}
	\captionsetup{width=1.0\textwidth, justification=justified}
	\label{tab:dataset_profile}
    \vspace{-2mm}
\end{table*}

\noindent\textbf{Scale and temporal characteristics.} We provide a comparative analysis of dataset properties in Tab.~\ref{tab:dataset_profile}, contrasting DynUAV with other established UAV-perspective MOT benchmarks. The statistics reveal a key design choice of our benchmark: while DynUAV may not have the highest sequence count, it features the longest average and minimum video durations. Unlike benchmarks composed of short clips, DynUAV prioritizes long-term temporal continuity, which is critical for real-world applications. In prolonged tracking scenarios, localization and identity errors tend to accumulate over time, resulting in catastrophic drift and track fragmentation. Consequently, the long-duration sequences in DynUAV present a more stringent and realistic challenge, specifically testing an algorithm's ability for long-term temporal modeling and drift suppression.

\begin{figure}[t!]
	\centering
	\includegraphics[width=0.48\textwidth]{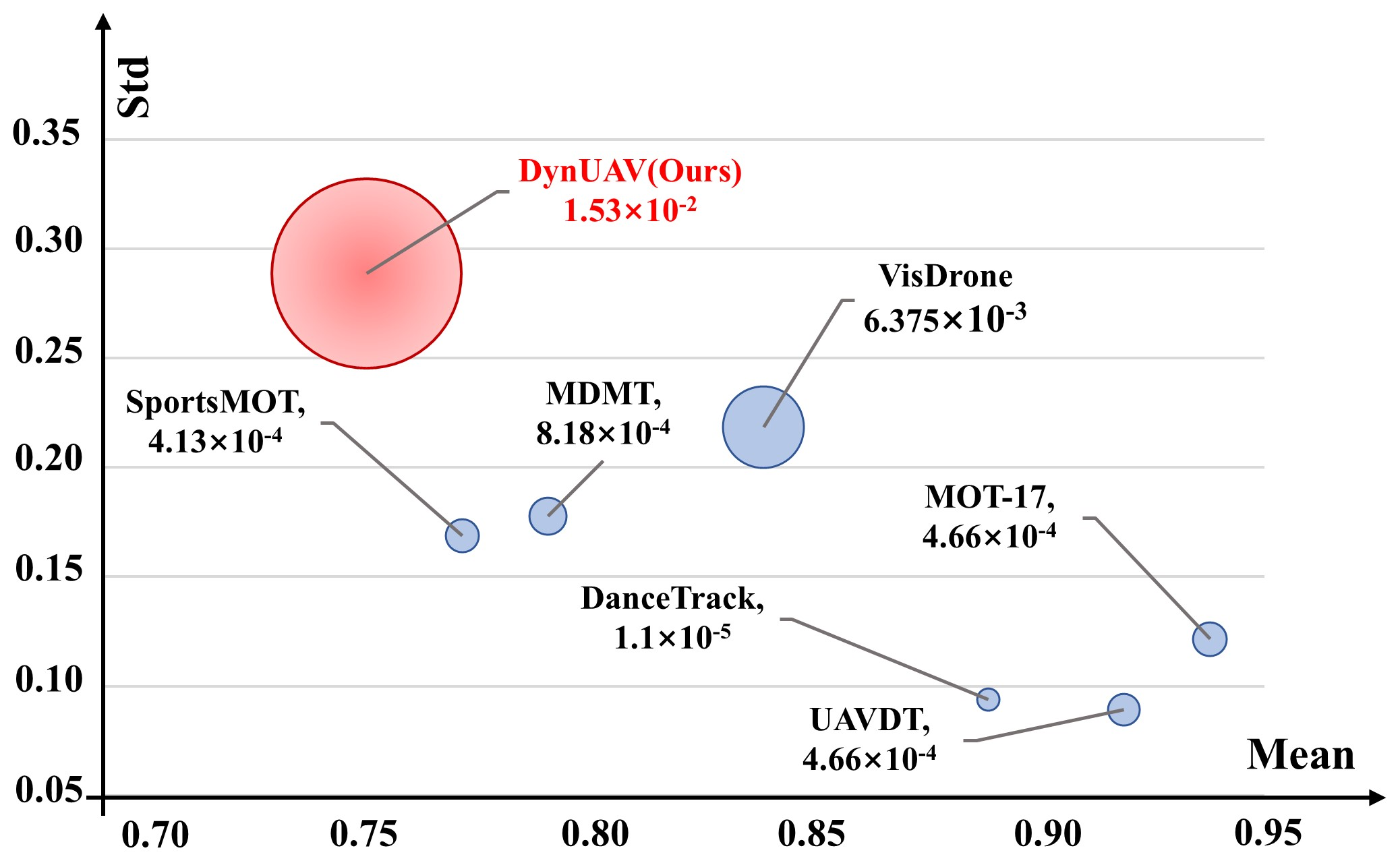} 
	\captionsetup{width=0.48\textwidth, justification=justified}
	\caption{Comparison of the mean and variance of IOU across datasets. The horizontal and vertical axes represent the mean and variance of IoU respectively. The bubble radius corresponds to the proportion of non-overlapping bounding boxes in adjacent frames.}
	\label{fig:IOU statistic}
\end{figure}

\noindent\textbf{IOU distribution and object scale characteristics.} Temporal consistency is measured using the Intersection-over-Union (IoU) between adjacent bounding boxes. To analyze temporal consistency, we compute the IoU between bounding boxes of the same object in adjacent frames. This metric effectively captures not only the magnitude of object motion but also \emph{apparent changes arising from variations in viewpoint, scale, and pose}. The bubble chart in Fig.\ref{fig:IOU statistic} reveals that DynUAV\textemdash which represented by the largest bubble size\textemdash occupies a unique ``low-mean, high-variance'' region. Its low mean IoU signifies large inter-frame displacements, while its high variance indicates the most diverse range of motion patterns. This stands in sharp contrast to benchmarks like MOT20\cite{dendorfer2020mot20}, which exhibits a more stable and predictable motion profile. Furthermore, to address the persistent challenge of small object tracking, we calculate the average object area relative to the frame area. As shown in Tab.\ref{tab:area statistic}, DynUAV ranks second only to MDMT\cite{liu2023robust} in terms of the smallest average object size, confirming that it poses a significant challenge for small object detection.

\begin{table}[t]
	\setlength{\tabcolsep}{4pt} 
	\centering
    \caption{Average bounding box area ratio across datasets.}
	\begin{tabular}{lcc}
		\toprule
		Dataset & \textit{avg.} \\
		\midrule
		MOT-17\cite{milan2016mot16} & \(1.3511\times10^{-2}\) \\
		MOT-20\cite{dendorfer2020mot20} & \(5.8318\times10^{-3}\) \\
		VisDrone\cite{zhu2021detection} & \(2.5820\times10^{-3}\) \\
		UAVDT\cite{du2018unmanned} & \(2.5922\times10^{-3}\) \\
		MDMT\cite{liu2023robust} & \(9.7198\times10^{-4}\) \\
		DanceTrack\cite{sun2022dancetrack} & \(3.1923\times10^{-2}\) \\
		\hline
		\parbox{2cm}{\textbf{DynUAV}\\(OURS)}  & \(1.1720\times10^{-3}\) \\
		\bottomrule
	\end{tabular}
	\captionsetup{width=0.3\textwidth, justification=justified}
	\label{tab:area statistic}
\end{table}

\begin{figure}[t!]
	\centering
	\includegraphics[width=0.5\textwidth]{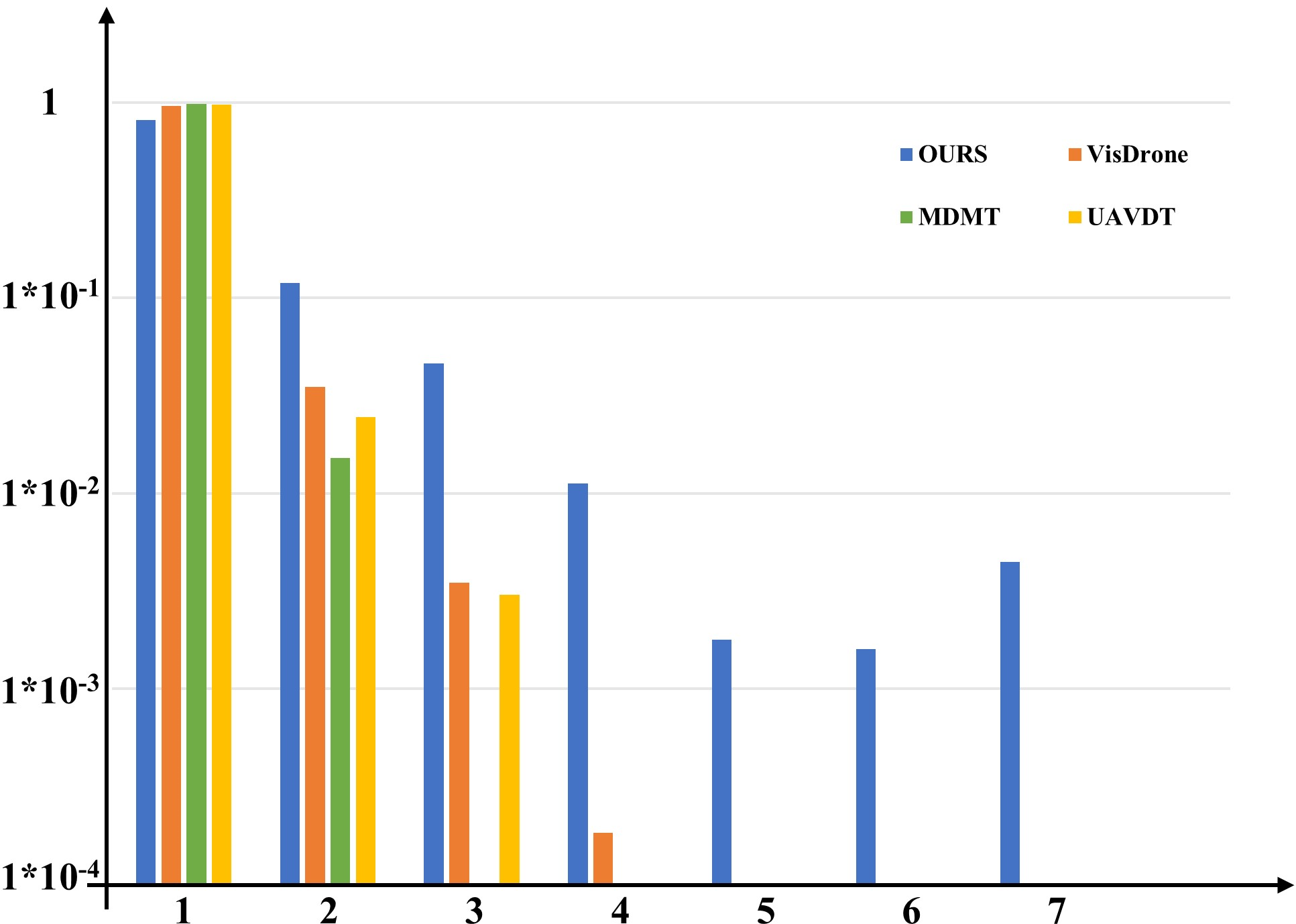} 
	\captionsetup{width=0.48\textwidth, justification=justified}
	\caption{Bar charts comparing the distributions of trajectory discontinuities across different datasets. Each bar chart displays the proportion of target IDs relative to the number of continuous trajectory segments.}
	\label{fig:trajectory Distribution}
\end{figure}

\noindent\textbf{Trajectory continuity and occlusion.} Trajectory fragmentation is a critical indicator of tracking difficulty. In the context of UAVs, it arises from two primary sources: (1) occlusions by other objects within the scene, and (2) targets being temporarily driven out of the field-of-view by the UAV's own complex maneuvers. Upon re-entry, these targets often undergo very significant changes in scale and viewpoint, presenting a considerable challenge for subsequent re-identification and trajectory re-linking. Fig.~\ref{fig:trajectory Distribution} visualizes the distribution of target instances according to their number of track fragments. While a vast majority of targets in VisDrone\cite{zhu2021detection}, UAVDT\cite{du2018unmanned}, and MDMT\cite{liu2023robust} are represented by a single, unbroken trajectory, DynUAV exhibits a significant long-tail distribution, with a substantial portion of all targets fragmented into multiple segments. This evidence unequivocally demonstrates that DynUAV is the \emph{most challenging benchmark when it comes to maintaining true long-term trajectory continuity.}

\section{Experiment}
\label{sec:experiment}

To validate the challenges introduced by DynUAV, we conduct a comprehensive evaluation of representative state-of-the-art MOT methods. First, we assess pretrained models on DynUAV to evaluate their generalization to UAV scenarios. Then, we fine-tune and re-evaluate these models to measure their adaptability. All reported results are based on the fine-tuned models.

\subsection{Experimental setup}
\label{subsec:Implement detail}

\noindent\textbf{Overview of evaluated algorithms.} To ensure a comprehensive evaluation, we benchmark DynUAV against 11 representative MOT algorithms, categorizing them into several families based on their core technical contributions:(1) unsupervised MOT algorithms (e.g., Path-Consistency\cite{lu2024self}), (2) motion model enhanced methods (e.g., OC-SORT\cite{maggiolino2023deep}, DiffMOT\cite{lv2024diffmot}), (3) single-stage association approaches (e.g., BoostTrack\cite{stanojevic2024boosttrack}, TrackTrack\cite{shim2025focusing}), (4) uncertainty-aware models (e.g., U2MOT\cite{liu2023uncertainty}) and (5) methods with adaptive memory or fusion mechanisms (e.g., AdapTrack\cite{shim2024adaptrack}, StrongSORT\cite{du2023strongsort}). This categorization enables us to analyze which algorithmic strategies are most resilient to the challenges presented by our dynamic UAV-perspective benchmark.

\noindent\textbf{Evaluation metrics.} We employ a comprehensive suite of established MOT metrics to assess tracker performance. The primary metric for overall accuracy is the Multi-Object Tracking Accuracy (MOTA\cite{bernardin2008evaluating}), which aggregates False Positives (FP), False Negatives (FN), and Identity Switches (IDSW). To specifically evaluate temporal identity consistency, we utilize the Identification F1-Score (IDF1\cite{ristani2016performance}). Furthermore, to decouple detection quality from association performance, we adopt the Higher Order Tracking Accuracy (HOTA\cite{luiten10higher}) metric, along with its decomposed components: Detection Accuracy (DetA) and Association Accuracy (AssA). This multifaceted evaluation allows for a nuanced analysis of whether performance bottlenecks originate from detection or association failures.

\noindent\textbf{Implementation details.}  All evaluated trackers were implemented following the default settings from their original publications or official codebases to ensure a fair comparison. We adopted YOLOv11\cite{khanam2024yolov11} as the unified detection backbone, training it for 100 epochs with an input resolution of 1280$\times$1280 on a single NVIDIA RTX 4090 GPU. The output detections then served as the input for all trackers. For purely detection-based methods (e.g., ByteTrack\cite{zhang2022bytetrack}, OC-SORT\cite{cao2023observation}), no Re-Identification (ReID) features were employed. For appearance-based trackers, we utilized two standard pretrained ReID models: StrongSORT\cite{du2023strongsort}, BoT-SORT\cite{aharon2022bot}, and Deep OC-SORT\cite{maggiolino2023deep} used OSNet\cite{jin2023identification} pretrained on MSMT17\cite{lv2018unsupervised}, while FastReID\cite{he2023fastreid}\textemdash trained on MOT17\cite{milan2016mot16}\textemdash was leveraged for TrackTrack\cite{shim2025focusing} and AdapTrack\cite{shim2024adaptrack}.

\subsection{Benchmark Analysis}

We analyze how tracking paradigms address core benchmark challenges. The overall results of 11 MOT trackers on DynUAV are presented in Tab.~\ref{tab:algo overall perforamce}.

\begin{table*}[t!]
	\centering
	\caption{Overall quantitative performance of representative MOT
		trackers on the DynUAV dataset.'↑'/'↓' indicates higher/lower values are better,
		respectively. Bold numbers are superior results.
		Note: All metrics, except for IDSW, are reported as percentages (e.g., 63.54 means 63.54\%).}    
	\begin{tabular}{l|cccccccccc}
		\toprule
		Algorithm & FP↓ & FN↓ & DetA↑ & MOTA↑ & HOTA↑ & IDF1↑ & AssA↑ & IDSW↓ & IDs↓ \\
		\midrule
		BoostTrack\cite{stanojevic2024boosttrack} & 19097 & 94087 & 49.52 & 56.72 & 53.71 & 63.77 & 58.98 & 609 & 1696 \\
		BoT-SORT\cite{aharon2022bot} & 17724 & 68090 & 57.84 & 66.88 & 59.86 & 68.81 & 62.65 & 1274 & 2264 \\
		ByteTrack\cite{zhang2022bytetrack} & 18935 & 75152 & 54.90 & 63.03 & 53.59 & 60.78 & 53.08 & 3136 & 4025 \\
		OC-SORT\cite{cao2023observation} & 18085 & 83611 & 53.40 & 60.69 & 51.73 & 58.12 & 50.81 & 1660 & 2738 \\
		Deep OC-SORT\cite{maggiolino2023deep} & 19799 & 67890 & 57.58 & 66.44 & 61.09 & 72.25 & 65.49 & 567 & 1598 \\
		StrongSORT\cite{du2023strongsort} & 24965 & 57302 & 59.38 & \textbf{68.18} & 60.87 & 71.21 & 63.15 & 1394 & 1749 \\
		DiffMOT\cite{lv2024diffmot} & \textbf{12273} & 68121 & 57.58 & 67.32 & 61.21 & 71.72 & 65.73 & 430 & 1272 \\
		Path-Consistency\cite{lu2024self} & 38596 & \textbf{46254} & \textbf{59.61} & 67.42 & 60.11 & 69.91 & 61.35 & 819 & 1411 \\
		AdapTrack\cite{shim2024adaptrack} & 21489 & 62181 & 58.84 & 67.96 & 62.33 & 73.26 & 66.71 & 583 & 1564 \\
		TrackTrack\cite{shim2025focusing} & 15617 & 71034 & 57.65 & 66.95 & \textbf{62.74} & \textbf{74.81} & \textbf{68.89} & \textbf{256} & \textbf{1125} \\
		U2MOT\cite{liu2023uncertainty} & 20087 & 91359 & 49.97 & 55.16 & 51.47 & 58.92 & 53.90 & 6452 & 8881 \\
		\bottomrule
	\end{tabular}
	\captionsetup{width=0.95\textwidth, justification=justified}
    \vspace{-1mm}
	\label{tab:algo overall perforamce}
    \vspace{-5mm}
\end{table*}

\noindent\textbf{Unsupervised learning.} The unsupervised Path-Consistency\cite{lu2024self} method achieves competitive detection scores (e.g., DetA), highlighting its ability to learn robust representations without relying heavily on annotations. However, its performance in association metrics (e.g., AssA, IDF1) is less impressive. This disparity suggests that \emph{its path consistency constraint, while effective for learning generalizable features, may not fully address the domain-specific requirement for highly discriminative appearance modeling necessary to resolve ambiguities in long-term UAV tracking.}

\noindent\textbf{Advanced motion models.} Trackers in this category are specifically designed to handle complex motion dynamics. OC-SORT\cite{cao2023observation}, with its observation-centric update mechanism, demonstrates a strong resilience to nonlinear motion and brief occlusions. Building upon this foundation, Deep OC-SORT\cite{maggiolino2023deep} achieves a significant performance leap by integrating appearance information, proving crucial for maintaining identity consistency amidst the severe viewpoint changes prevalent in DynUAV. Notably, DiffMOT\cite{lv2024diffmot} attains a comparable top-tier performance, showcasing the impressive effectiveness of its diffusion-based motion prior in modeling the complex and highly stochastic motion patterns inherent to truly agile UAV flight.

\noindent\textbf{Adaptive fusion mechanisms.} StrongSORT\cite{du2023strongsort}, a comprehensive enhancement to the SORT series, integrates both Camera Motion Compensation (CMC) and an advanced fusion of appearance and motion cues. On DynUAV, it achieves a high detection-related score (DetA), yet its core association metrics (AssA, IDF1) notably lag behind top performers like Deep OC-SORT\cite{maggiolino2023deep}. This performance gap suggests a potential vulnerability: its complex fusion strategy may over-rely on appearance features, which are frequently degraded by the severe motion blur and rapid pose variations ever-present in our benchmark. This highlights a critical trade-off between multi-cue fusion complexity and resilience to the dynamic observational noise induced by severe camera ego-motion.

\noindent\textbf{Adaptive memory methods.} AdapTrack\cite{shim2024adaptrack} is specifically designed for long-term tracking consistency via a dynamic trajectory state update mechanism. Given that DynUAV has the longest average sequence duration among its peers, this inherent characteristic of our benchmark clearly benefits AdapTrack\cite{shim2024adaptrack} by enabling its adaptive memory mechanism to fully demonstrate its advantage in managing long-term target dynamics and appearance variations. Consequently, it achieves a competitive performance, particularly in maintaining tracklet integrity across extended intervals.

\noindent\textbf{Uncertainty-aware matching.} U2MOT\cite{liu2023uncertainty} incorporates uncertainty modeling to dynamically calibrate detection and ReID confidences. Theoretically, this capability should provide a distinct advantage on DynUAV, where rapid UAV motion frequently results in blurry and low-resolution targets. However, its overall performance suggests that the severe appearance variations and unpredictable motion in our benchmark may pose challenges that exceed the limits of its current uncertainty estimation model.

\noindent\textbf{Single-stage association.} Among single-stage trackers, TrackTrack\cite{shim2025focusing} achieves state-of-the-art performance, leading in multiple association metrics, including IDF1, AssA and HOTA. This underscores the robustness of its unified matching process, which holistically leverages detections across all confidence levels. In stark contrast, the strategy employed by BoostTrack\cite{stanojevic2024boosttrack} proves less effective on DynUAV. Its aggressive approach of amplifying thresholds and generating pseudo-boxes, while designed to \emph{recover missed detections, appears to backfire in our cluttered and dynamic scenes, introducing excessive FPs and causing severe identity fragmentation}.

\noindent\textbf{Summary.} 
Results on DynUAV demonstrate that trackers employing robust motion models (e.g., OC-SORT\cite{cao2023observation}, DiffMOT\cite{lv2024diffmot}) combined with lightweight appearance cues (e.g., Deep OC-SORT\cite{maggiolino2023deep}), or those specifically designed for uncertain scenarios (e.g., AdapTrack\cite{shim2024adaptrack}, U2MOT\cite{liu2023uncertainty}, TrackTrack\cite{shim2025focusing}), perform best in addressing real-world UAV challenges. Therefore, DynUAV serves as a vital benchmark for developing accurate and reliable MOT methods under dynamic aerial perspectives.

\subsection{Cross-Dataset Comparative Analysis}

\begin{table*}
	\centering
    \caption{Performance Degradation: DynUAV vs MOT-17\cite{milan2016mot16}, MOT-20\cite{dendorfer2020mot20} and DanceTrack\cite{sun2022dancetrack}. Performance drop relative to conventional high-frequency test sets. Negative values indicate degradation, with the largest decrease in each metric highlighted in bold.}
	\begin{tabular}{l|ccc|ccc|cccc}
		\toprule
		\multirow{2}{*}{Algorithm} & \multicolumn{3}{c}{MOT-17} & \multicolumn{3}{c}{MOT-20} & \multicolumn{3}{c}{DanceTrack} \\
		\cline{2-10}
		& MOTA↑ & IDF1↑ & HOTA↑ & MOTA↑ & IDF1↑ & HOTA↑ & MOTA↑ & IDF1↑ & HOTA↑ \\
		\midrule
		BoostTrack\cite{stanojevic2024boosttrack} & -23.98 & -18.43 & \textbf{-12.89} & -20.98 & \textbf{-18.23} & \textbf{-12.69} & - & - & - \\
		BoT-SORT\cite{aharon2022bot} & -13.72 & -10.69 & -4.74 & -10.82 & -7.49 & -2.74 & - & - & - \\
		ByteTrack\cite{zhang2022bytetrack} & -17.27 & -16.52 & -9.51 & -14.77 & -14.42 & -7.71 & - & - & - \\
		OC-SORT\cite{cao2023observation} & -17.31 & -19.38 & -11.47 & -14.82 & -17.78 & -10.37 & \textbf{-31.51} & 3.22 & -3.37 \\
		Deep OC-SORT\cite{maggiolino2023deep} & -12.96 & -8.35 & -3.81 & -9.16 & -6.95 & -2.81 & -25.86 & 10.75 & -0.21 \\
		StrongSORT\cite{du2023strongsort} & -11.42 & -8.29 & -3.53 & -5.62 & -5.79 & -1.73 & -22.92 & 16.01 & 5.27 \\
		DiffMOT\cite{lv2024diffmot} & -12.48 & -7.58 & -3.29 & - & - & - & -25.48 & 8.72 & -1.09 \\
		Path-Consistency\cite{lu2024self} & -13.48 & -9.69 & -4.89 & - & - & - & -  & - & - \\
		AdapTrack\cite{shim2024adaptrack} & -11.94 & -9.04 & -3.37 & -7.04 & -7.44 & -2.67 & - & - & - \\
		TrackTrack\cite{shim2025focusing} & -14.85 & -8.29 & -4.36 & -11.05 & -6.09 & -2.96 & -26.65 & 7.01 & \textbf{-3.76} \\
		U2MOT\cite{liu2023uncertainty} & \textbf{-24.54} & \textbf{-19.28} & -12.73 & \textbf{-21.94} & -17.28 & -11.23 & - & - & - \\
		\bottomrule
	\end{tabular}
	\captionsetup{width=1.0\textwidth, justification=justified}
	\label{tab:cross dataset per compar}
    \vspace{-1mm}
\end{table*}

We compare all 11 models across DynUAV, MOT17\cite{milan2016mot16}, MOT20\cite{dendorfer2020mot20}, and DanceTrack\cite{sun2022dancetrack} in Tab.~\ref{tab:cross dataset per compar}.

\noindent\textbf{vs. MOT17.} In contrast to MOT17\cite{milan2016mot16}, trackers on DynUAV exhibit the most significant performance drops in MOTA and IDF1. The decline in MOTA, a metric highly sensitive to False Negatives (FN), is primarily caused by frequent detection failures. These failures stem from the rapid camera ego-motion and drastic scale changes in DynUAV, which are absent in MOT17's smoother camera work. Similarly, the sharp decrease in IDF1 is attributed to frequent IDSW and track fragmentations, triggered by the extreme viewpoint variations unique to our benchmark. This comparison demonstrates that DynUAV shifts the core challenge from \emph{occlusion in dense crowds to maintaining consistency under severe observational dynamics.}

\noindent\textbf{vs. MOT20.} The comparison with MOT20\cite{dendorfer2020mot20} further underscores the unique challenges of DynUAV, as trackers consistently underperform across all key metrics (MOTA, IDF1, and HOTA). This universal performance degradation confirms that DynUAV presents a \emph{more formidable overall tracking challenge}. While MOT20 primarily tests an algorithm's ability to handle density-induced ambiguity in extremely crowded scenes, the difficulties in DynUAV stem from motion-induced instabilities: detection failures from severe ego-motion (harming MOTA), identity disruptions from drastic viewpoint shifts (harming IDF1), and the compounded impact on HOTA. Therefore, MOT20 and DynUAV represent distinct and complementary dimensions of difficulty: crowd-induced ambiguity versus motion-induced instability.

\noindent\textbf{vs. DanceTrack.} The comparison with DanceTrack\cite{sun2022dancetrack} reveals a key distinction: trackers on DynUAV suffer a far more significant drop in MOTA. This is because DanceTrack, with its static camera and clean backgrounds, provides a near-perfect detection environment where low FN/FP rates make MOTA predominantly a measure of association quality (IDSW). In stark contrast, the severe ego-motion in DynUAV introduces substantial detection instabilities due to drastic viewpoint and scale variations, fundamentally degrading the MOTA score. This underscores how DynUAV fills a critical gap by evaluating tracker robustness under severe camera motion. The resulting phenomenon\textemdash a moderate HOTA score coupled with a drastically reduced MOTA\textemdash indicates that the primary difficulty of our benchmark stems not from localization ambiguity, but from the systemic stress placed on the entire tracking pipeline, a crucial characteristic for evaluating real-world robotic systems.

\subsection{Qualitative Analysis and Visualization}

\begin{figure}
	\centering
	\includegraphics[width=0.5\textwidth]{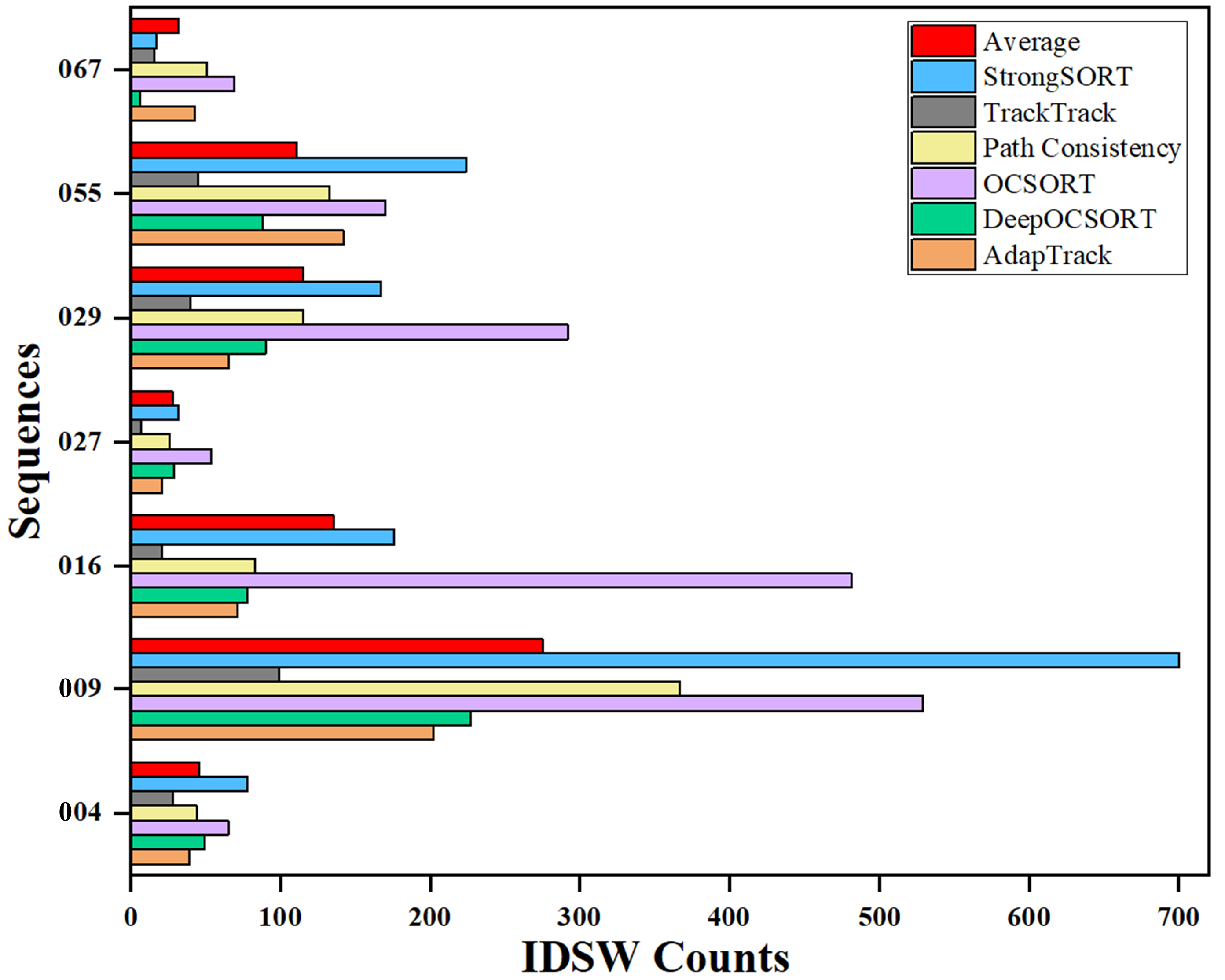}
	\captionsetup{width=0.47\textwidth, justification=justified}
	\caption{Fine-grained IDSW analysis. IDSW counts per algorithm across test sequences in DynUAV, revealing performance fluctuations in diverse scenarios.}
	\label{fig: IDSW bar chat}
\end{figure}

Our qualitative analysis explores the root causes of tracking failures by correlating them with specific scene attributes. Fig.~\ref{fig: IDSW bar chat} clearly identifies overpass (Seq. 009) and campus entrance (Seq. 016) as the most challenging scenarios. These sequences combine complex camera motions with persistent occlusions and small objects, creating a perfect storm for identity discontinuity. In contrast, sequences with sparser traffic (027) or initially occluded targets that were not annotated (067) exhibited significantly fewer IDSW. 

This pattern is further reflected in individual tracker performance. For instance, StrongSORT\cite{du2023strongsort} struggles most in the highly dynamic Sequence 009, while the long-term memory mechanism of AdapTrack\cite{shim2024adaptrack} provides greater consistency. The results reveal two main causes: (1) inability to re-identify targets after long-term occlusion, and (2) identity fragmentation due to drastic viewpoint changes. These findings confirm that \emph{maintaining identity consistency under severe ego-motion remains a critical, unsolved challenge for current MOT methods\textemdash one that DynUAV is specifically designed to expose and catalyze research on.}


\subsection{Ablation Studies and Analysis} 

\begin{table*}[t!]
	\centering
    \caption{Comparison of MOT performance with and without CMC. "w" and "o" denote with and without CMC, respectively.}
	\begin{tabular}{l|c|cccccccccc}
		\toprule
		\multicolumn{2}{c}{Algorithm} & FP↓ & FN↓ & DetA↑ & MOTA↑ & HOTA↑ & IDF1↑ & AssA↑ & IDSW↓ & IDs↓\\
		\midrule
		\multirow{2}{*}{AdapTrack\cite{shim2024adaptrack}} & o & 19902 & 74779 & 55.72 & 63.57 & 56.99 & 65.53 & 58.97 & 1123 & 2035 \\
		\cline{2-11}
		& w & 21489 & 62181 & 58.84 & 67.96 & 62.33 & 73.26 & 66.71 & 583 & 1564 \\
		\hline
		\multirow{2}{*}{TrackTrack\cite{shim2025focusing}} & o & 15030 & 75740 & 56.30 & 65.31 & 60.64 & 71.04 & 65.95 & 439 & 1343 \\
		\cline{2-11}
		& w & 15617 & 71034 & 57.65 & 66.95 & 62.74 & 74.81 & 68.89 & 256 & 1125 \\
		\hline
		\multirow{2}{*}{U2MOT\cite{liu2023uncertainty}} & o & 19265 & 92843 & 49.42 & 54.81 & 50.95 & 58.85 & 53.34 & 6731 & 8919 \\
		\cline{2-11}
		& w & 20087 & 91359 & 49.97 & 55.16 & 51.47 & 58.92 & 53.90 & 6452 & 8881 \\
		\bottomrule
	\end{tabular}
	\captionsetup{width=1.0\textwidth, justification=justified}
	\label{tab:CMC comparasion}
    \vspace{-1mm}
\end{table*}

\noindent\textbf{Quantitative impact of CMC.} To quantitatively isolate the impact of ego-motion, a core challenge of DynUAV, we conducted an ablation study on CMC. We generated a unified set of CMC parameters for our benchmark and systematically evaluated several trackers by selectively enabling or disabling this module. As shown in Tab.~\ref{tab:CMC comparasion}, enabling CMC unequivocally improves association performance\textemdash substantially reducing IDSW and boosting AssA far more than DetA. This confirms its primary role in stabilizing motion prediction by decoupling target movement from camera-induced jitter. However, we identified a nuanced trade-off: the image warping inherent to CMC can introduce artifacts, occasionally increasing FP. This insight reveals that while CMC is crucial for robust association, its naive application can be suboptimal for detection. Developing more advanced, jointly optimized detection-and-compensation frameworks therefore emerges as a promising future research direction.

\begin{figure}
	\centering
	\includegraphics[width=0.5\textwidth]{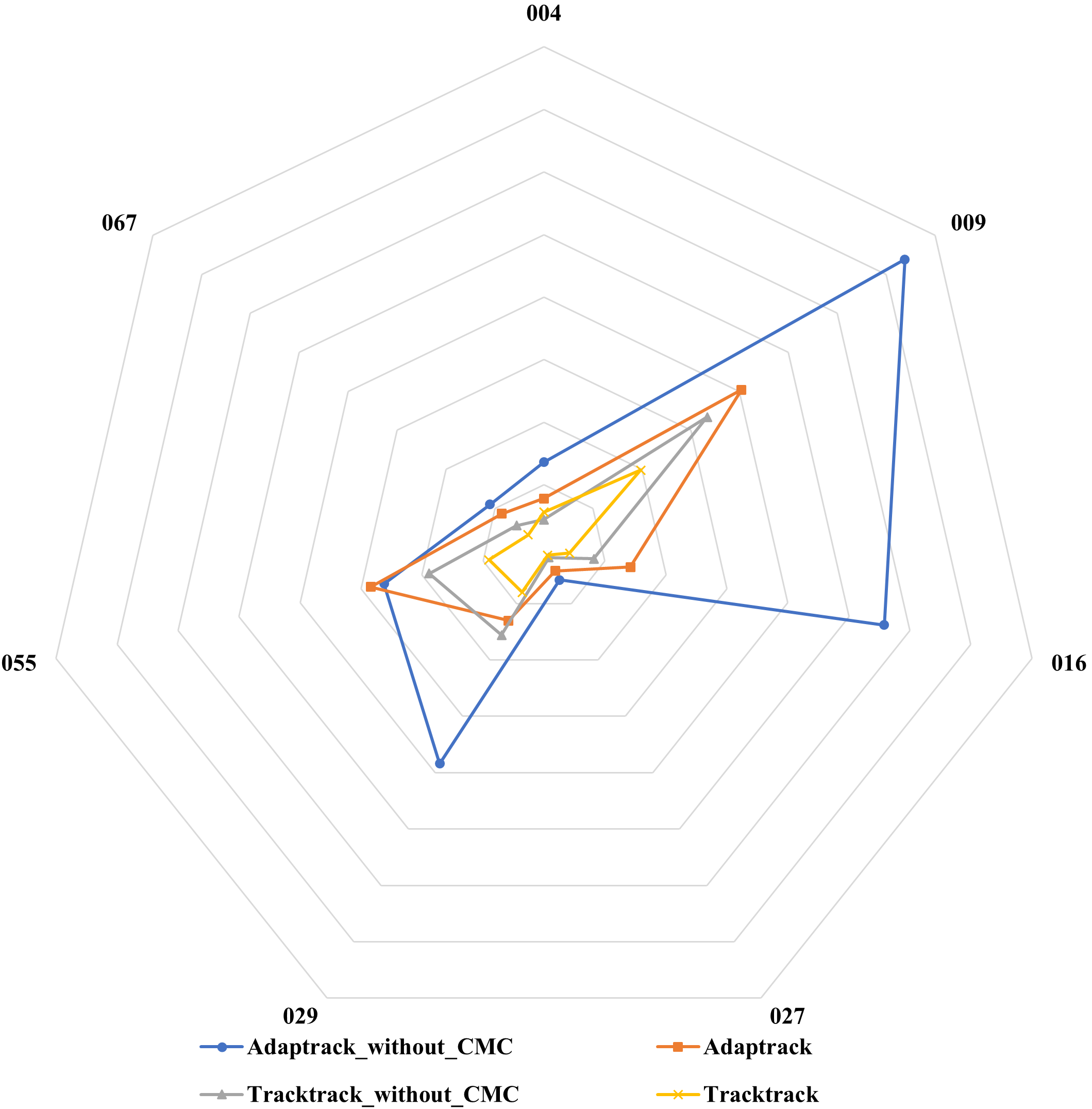}
	\captionsetup{width=0.47\textwidth, justification=justified}
	\caption{Per-Sequence IDSW with CMC. A radar plot showing the number of IDSW for each color-coded algorithm across all test sequences.}
	\label{fig:IDSW visualize in CMC}
    \vspace{0mm}
\end{figure}

\noindent\textbf{Fine-grained analysis of CMC across scenarios.} We leverage CMC as an analytical tool to diagnose the impact of ego-motion across different scenarios. The radar chart in Fig.~\ref{fig:IDSW visualize in CMC} reveals a clear pattern: the more severe the camera motion within a sequence, the greater the performance gain afforded by CMC. This is exemplified in sequences like Sequence 009, characterized by frequent panning, and Sequence 029 (construction site), which features complex camera trajectories. In these cases, severe ego-motion is intertwined with perceptual challenges like occlusion and small object scales, and it is precisely in such scenarios that we observe the most significant reduction in IDSW for both AdapTrack\cite{shim2024adaptrack} and TrackTrack\cite{shim2025focusing}. Conversely, the benefits of CMC are marginal in sequences with milder camera dynamics. This contrast substantiates our central claim: the diverse UAV dynamics are a primary source of tracking difficulty in DynUAV, and the magnitude of CMC's improvement serves as a direct proxy for the intensity of the ego-motion challenge inherent in each sequence.
\section{Discussion}

Building upon the experimental findings introduced by DynUAV\textemdash particularly in handling rapid camera motion and small object tracking\textemdash we argue that future MOT research should move beyond assumptions of static or smooth-motion. The performance degradation observed across methods, combined with the improvements gained from motion compensation, highlights the need for models capable of reasoning under highly dynamic UAV conditions. Promising directions include integrating multi-modal sensors to enhance small object localization, augmenting trajectory annotations to support behavioral prediction, and developing language-in-the-loop tracking interfaces for UAV navigation. DynUAV thus provides a realistic and challenging benchmark to foster such advancements.

\section{Conclusion}

This paper introduces DynUAV, a UAV-perspective MOT benchmark designed to challenge the smooth-motion assumption prevalent in current methods. By systematically incorporating agile UAV maneuvers, DynUAV introduces severe ego-motion, leading to complex apparent trajectories, drastic scale changes, and motion blur. Our comprehensive evaluations demonstrate that these dynamic conditions reveal critical limitations of state-of-the-art trackers, whose performance is primarily bottlenecked by association failures and detection losses. These results underscore the urgent need for trackers that are robust to the severe observational dynamics inherent in real-world UAV flight.
{
    \small
    \bibliographystyle{ieeenat_fullname}
    \bibliography{main}

}


\end{document}